\documentclass[journal]{IEEEtran}
\usepackage{amsmath,amsfonts}
\usepackage{amssymb}
\usepackage{algorithmic}
\usepackage[ruled]{algorithm2e}
\usepackage{array}
\usepackage{balance}
\usepackage{booktabs}       
\usepackage[caption=false,font=normalsize,labelfont=sf,textfont=sf]{subfig}
\usepackage{color}
\usepackage{float}
\usepackage{graphicx}
\usepackage{hyperref}       
\usepackage[utf8]{inputenc} 
\usepackage{microtype}      
\usepackage{multirow}
\usepackage{nicefrac}       
\usepackage{textcomp}
\usepackage{stfloats}
\usepackage{url}
\usepackage{verbatim}
\usepackage{xcolor}         

\def\BibTeX{{\rm B\kern-.05em{\sc i\kern-.025em b}\kern-.08em
    T\kern-.1667em\lower.7ex\hbox{E}\kern-.125emX}}

\begin{document}
\title{Perturbation-based Self-supervised Attention\\ for Attention Bias in Text Classification}
\author{Huawen Feng, \quad 
        Zhenxi Lin, \quad
        Qianli Ma*, ~\IEEEmembership{Member, IEEE}
\thanks{Huawen Feng, Zhenxi Lin, and Qianli Ma are with the School
of Computer Science and Engineering, South China University of Technology,
Guangzhou, 510006, China (e-mail: 541119578@qq.com, 786450794@qq.com, qianlima@scut.edu.cn, *corresponding author)}}

\markboth{Journal of \LaTeX\ Class Files,~Vol.~14, No.~8, August~2021}%
{Shell \MakeLowercase{\textit{et al.}}: A Sample Article Using IEEEtran.cls for IEEE Journals}

\maketitle

\begin{abstract}
In text classification, the traditional attention mechanisms usually focus too much on frequent words, and need extensive labeled data in order to learn. This paper proposes a perturbation-based self-supervised attention approach to guide attention learning without any annotation overhead. Specifically, we add as much noise as possible to all the words in the sentence without changing their semantics and predictions. We hypothesize that words that tolerate more noise are less significant, and we can use this information to refine the attention distribution. Experimental results on three text classification tasks show that our approach can significantly improve the performance of current attention-based models, and is more effective than existing self-supervised methods. We also provide a visualization analysis to verify the effectiveness of our approach.
\end{abstract}


\section{Introduction}
\label{sec:intro}

\IEEEPARstart{A}{ttention} mechanisms~\cite{bahdanau2014neural,luong2015effective,Vaswani2017attention} play an essential role in Natural Language Processing (NLP) and have been shown to be effective in various text classification tasks, such as sentiment analysis~\cite{lin2017structured,tang2019progressive,choi2020lessismore}, document classification~\cite{yang2016hierarchical} and natural language inference~\cite{chen2017enhanced}.
They achieve significant performance gains, and can be used to provide insights into the inner workings of the model. Generally, the attention learning procedure is conditioned on access to large amounts of training data without additional supervision information.

Although the current attention mechanisms have achieved remarkable performance, several problems remain unsolved. 
First, learning a good attention distribution without spurious correlations for neural networks requires large volumes of informative labeled data~\cite{barrett2018sequence,bao2018deriving}. As described in the work of Wallace et al.~\cite{wallace2020concealed}, after inserting 50 poison examples with the name  ``\textit{James Bond}'' into its training set, a sentiment model will frequently predict a positive whenever the input contains this name, even though there is no correlation between the name and the prediction. 
Second, attention mechanisms are prone to focus on high-frequency words with sentiment polarities and assign relatively high weights to them~\cite{xu2018random,li2018transformation,tang2019progressive}, while the higher frequency does not imply greater importance. 

\begin{figure*}[t!]
	\centering
	\includegraphics[width=1.0\linewidth]{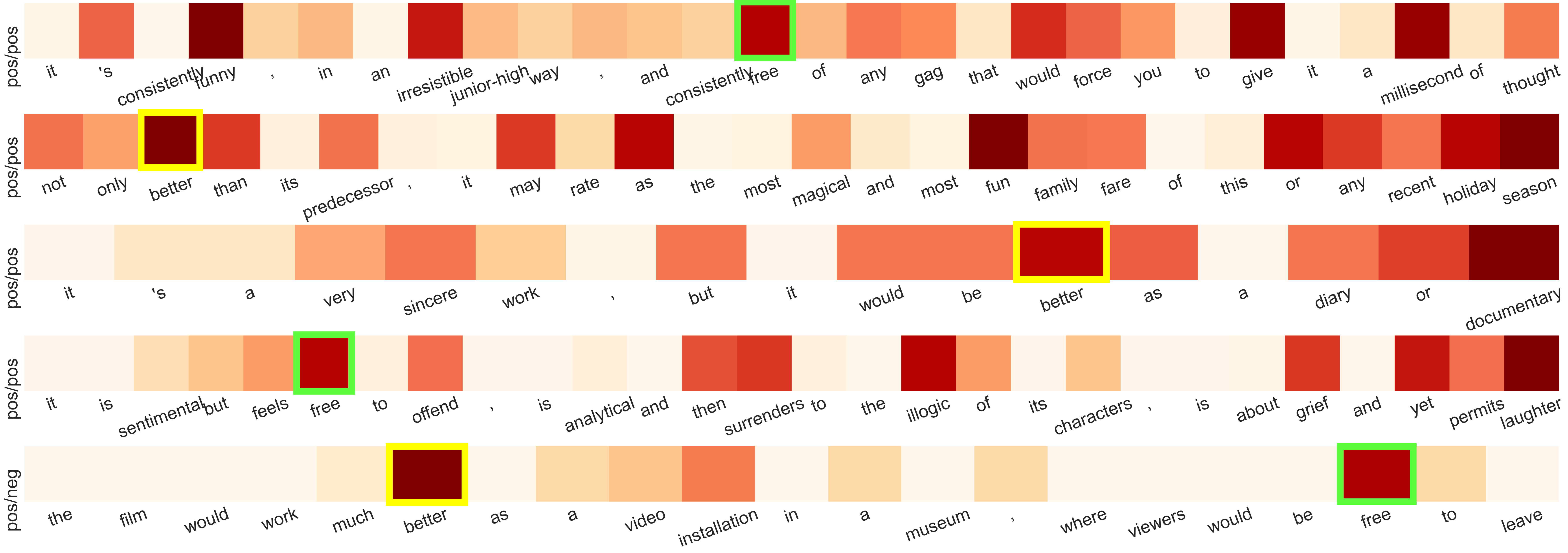}
    \caption{The attention visualization for five sentences. The "A/B" style tags before each row mean the model's prediction is A and the label is B. The first four sentences are selected from training sets as representatives containing high-frequency words - "better" (yellow box) and "free" (green box). The last sentence including both of the two words is selected from testing sets, typically showing that the distribution of attention weights when some words in the sentence appear frequently in the corpus but are unimportant to the current prediction.}
    \label{Fig:problem}
\end{figure*}

Especially when there’s an adversative relation in a text, some high-frequency words with strong sentiment valence need to be selectively ignored based on the context of the whole text. In these cases, these words will mislead the model because the important words don’t get enough attention. The sentences in Figure~\ref{Fig:problem} illustrate this problem.
In most training sentences, as shown in the first four rows, ``\textit{better}'' and ``\textit{free}'' appear with positive sentiment, which makes the attention mechanism accustomed to attaching great importance to them and relating them to positive predictions. However, the two words are used ironically in the fifth sentence, and the model pays the most attention to them while the critical word -- ``\textit{leave}'' -- is not attended to, resulting in an incorrect prediction. Based on these observations, there’s reason to believe that the attention mechanisms could be improved for text classification.

To tackle this problem the most direct solution is to add human supervision collected by manual annotation~\cite{zhang2016rationale,bao2018deriving,camburu2018snli} or special instruments~\cite{barrett2018sequence,sood2020improving,sood2020interpreting,malmaud2020bridging} (e.g., eye-tracking), to provide an inductive bias for attention. These approaches are costly, the labeling is entirely subjective, and there is often high variance between annotators.
In particular, Sen et al.~\cite{sen2020humanattention} point out that there is a huge difference between machine and human attention and it is difficult to map human attention to machine attention.

Another flexible solution is to measure attribution scores, i.e., how much each token in a text contributes to the final prediction, to approximate an importance distribution as an attention supervision signal~\cite{li2016understanding,choi2019counterfactual,tang2019progressive,choi2020lessismore}. Generally, the attribution scores are obtained by masking each token one by one to generate counterfactual examples, reflecting the difference in the softmax probability of the model after masking each token. These approaches have little or no additional annotation overhead and augment supervision information from the training corpus to refine the attention distribution. Despite their success,  masking schemes can give rise to an out-of-distribution (OOD) problem~\cite{hendrycks2016baseline,chang2018explaining,yi2020information}. That is, the generated counterfactuals deviate from the training data distribution of the target model, resulting in an overestimation of the contribution of unimportant tokens. The OOD problem induced by existing masking schemes makes it difficult to identify whether high-scoring tokens contribute significantly to the prediction. Furthermore, most of them are limited to generating uniform attention weights for the selected important words. Obviously, the contribution of different important words to the model should also be different according to the context, e.g., the word \textit{leave} should have a higher attention weight than \textit{better} and \textit{free} for the fifth sentence in Figure~\ref{Fig:problem}.

Some efforts reveal that the output of neural networks can be theoretically guaranteed to be invariant for a certain magnitude of input perturbations through establishing the concept of maximum safety radius~\cite{wu2020game,la2020assessing} or minimum disturbance rejection~\cite{weng2018evaluating}. In simple terms, these approaches evaluate the minimum distance of the nearest perturbed text in the embedding space that is classified differently from the original text. 
Inspired by this work, we propose a novel perturbation-based self-supervised attention learning method without any additional annotation overhead for text classification. Specifically, we design an attention supervision mining mechanism called Word-based Concurrent Perturbation (WBCP), which effectively calculates an explainable word-level importance distribution for the input text. Concretely, WBCP tries to concurrently add as much noise as possible to perturb each word embedding of the input, while ensuring that the semantics of input and the classification outcome is not changed. Under this condition, the words that tolerate more noise are less important and the ones sensitive to noise deserve more attention.
We can use the permissible perturbation amplitude as a measure of the importance of a word, where small amplitude indicates that minor perturbations of that word can have a significant influence on the semantic understanding of input text and easily lead to prediction error.

According to the inverse distribution of perturbation amplitude, we can get sample-specific attention supervision information. Later, we use this supervision information to refine the attention distribution of the target model and iteratively update it.
Notably, our method is model-agnostic and can be applied to any attention-based neural network. It generates attention supervision signals in a self-supervised manner to improve text classification performance without any manual labeling and incorporates Perturbation-based Self-supervised Attention (PBSA) to avoid the OOD problem caused by the masking scheme. In addition, it can also generate special attention supervision weights adaptively for each sample based on the perturbation amplitude, rather than allocate them uniformly.

In summary, the contributions of this paper are as follows:

(1) Through analysis of current methods, we point out the disadvantages and drawbacks of current attention mechanisms for text classification.

(2) We propose a simple yet effective approach to automatically mine the attribution scores for the input text, and use it as supervision information to guide the learning of attention weights of target models.

(3) We apply our approach to various text classification tasks, including sentence classification, document categorization, and aspect-level sentiment analysis. Extensive experiments and visualization analysis show the effectiveness of the proposed method in improving both model prediction accuracy and robustness.

(4) Theoretically, our algorithm can be applied to the models with attention mechanisms, but it is impossible to compare with all of them. Considering this, we conduct our experiments on several typical baselines (LSTM, BERT~\cite{devlin2018bert}, DEBERTA~\cite{he2020deberta}, ELECTRA~\cite{clark2020electra}, Memory Net~\cite{mn2016tang}, etc.) to justify the effectiveness of our method. Notably, we also compared our algorithm with other advanced attention self-supervised methods (PGAS~\cite{su2021enhanced}, AWAS~\cite{tang2019progressive}, SANA~\cite{choi2020lessismore}).

\section{Related work}
\begin{figure*}[h!]
	\centering
	\includegraphics[width=1.0\linewidth]{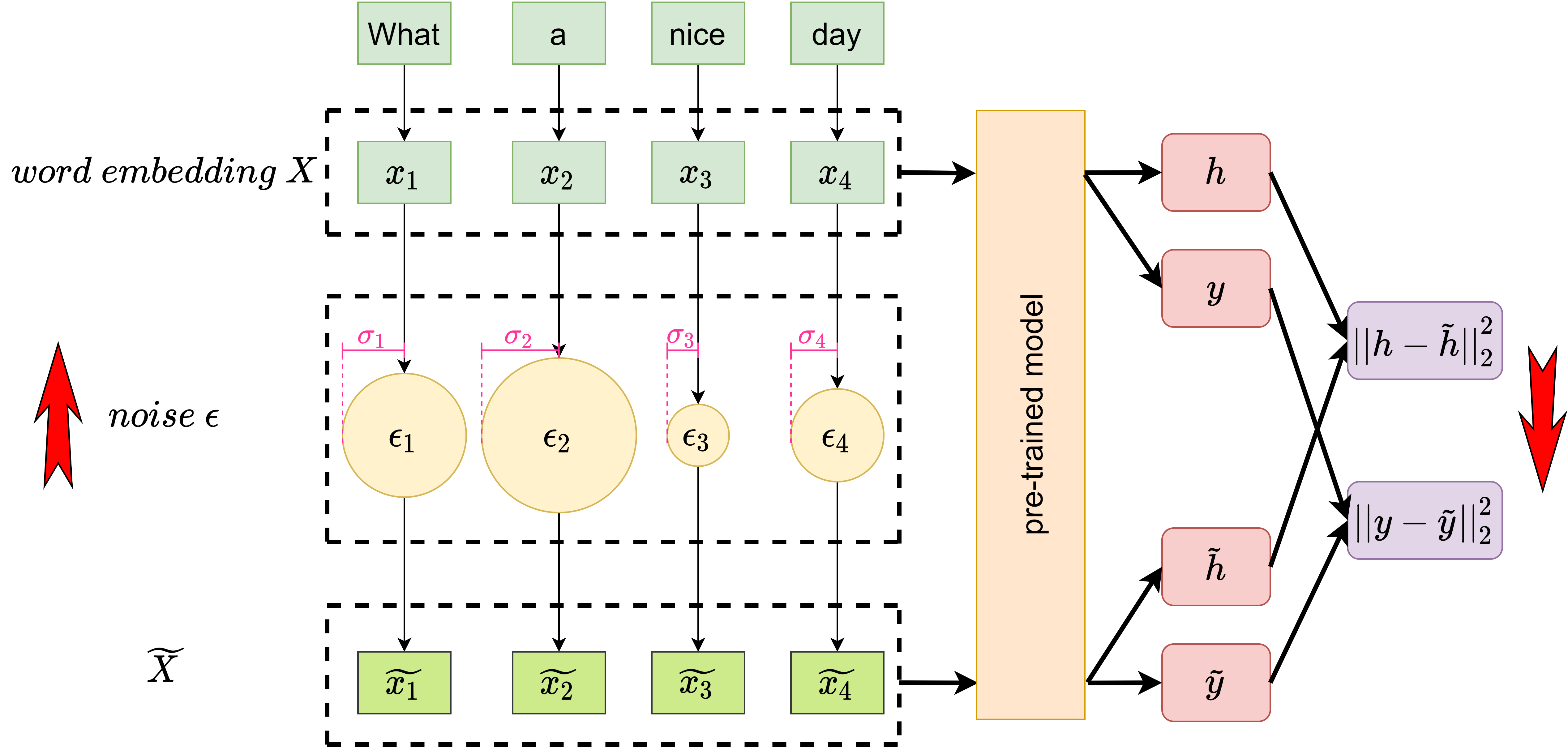}
    \caption{The diagram of WBCP. The left part of the figure corresponds to the last term of Eq.~(\ref{eq-rewrite}), which illustrates the process of adding noise that follows a Gaussian distribution to each word. The right part of the figure corresponds to the first two terms of Eq.~(\ref{eq-rewrite}), indicating the constraint of trying to not change the semantics and predictions after the noise is introduced. 
    }
    \label{Fig:perturb}
\end{figure*}

Work related to our method can be categorized into three types: Introducing human attention; using external resources or tools; and using self-supervision.

\textbf{Introducing human attention}\quad 
Adding human supervision to attention has been shown to effectively alleviate attention bias and improve model prediction accuracy on a range of tasks~\cite{zhang2016rationale,camburu2018snli,sood2020improving,sood2020interpreting,malmaud2020bridging}. In general, the annotators need to explicitly highlight the important words or rationales~\cite{zhang2016rationale,bao2018deriving,camburu2018snli} for the given sample. Obviously, the annotation is very labor-intensive and expensive in real-world scenarios, so an alternative is to use implicit signals such as eye gaze~\cite{barrett2018sequence,sood2020improving,sood2020interpreting,malmaud2020bridging}. For these methods, it is expected that the model can generate similar attention to human supervision. However, human recognition and model reasoning processes may be inconsistent~\cite{jacovi2020towards}, and aligning the two is challenging~\cite{sen2020humanattention}.

\textbf{Using external resources or tools}\quad
With the development of NLP, many corpora and tools, such as Dependency Tree and Synonym Dictionary, are created to obtain a deeper understanding of words and sentences. Therefore, some methods~\cite{kamigaito2017supervised,zou2018lexicon,nguyen2018killed,zhao2020attention} that generate attention supervision information according to existing corpora and tools emerge. For example, Nguyen et al.~\cite{nguyen2018killed} introduce attention supervision information based on important words selected by semantic word lists and dependency trees. Similarly, Zhao et al.~\cite{zhao2020attention} first train the model on the document-level sentiment classification and then transfer the attention knowledge to a fine-grained one for aspect-level sentiment classification. And Hu et al.~\cite{hu2023multi} introduce the tree structure's representation into attention computations. However, these methods still rely on annotations based on parsers or external resources, and the performance depends heavily on the quality of the parser.

\textbf{Self-supervised attention learning}\quad
Currently, self-supervised attention learning frameworks~\cite{li2016understanding,choi2019counterfactual,tang2019progressive,choi2020lessismore,su2021enhanced} have become the mainstream method because they do not require additional annotation overhead. 
They usually mask or erase each token one by one and quantify the difference in predictions of the model after masking each token, to approximate an importance distribution as attention supervision information. For example, Tang et al.~\cite{tang2019progressive} divide the words in sentences into the active set and the misleading set by progressively masking each word with respect to the maximum attention weight, and augment them to make the model focus on the active context words.
Similarly, Choi et al.~\cite{choi2020lessismore} adopt the masking method to find the unimportant words and gradually reduce their weights. These methods use a self-supervised paradigm to mine important words, which can greatly reduce the annotation cost and improve the robustness of the model. Nevertheless, the masking scheme they follow has an OOD problem. The counterfactuals generated by the mask operation deviate from the original training set distribution, which easily leads to the over-evaluation of unimportant words. In addition, the above methods usually assign the same weight to the extracted important words, but in our opinion, different words should have different contributions to the classification.

\section{Proposed method}
In this section, we propose a Perturbation-based Self-supervised Attention (PBSA) mechanism to enhance the attention learning process and provide a good inductive bias. We first design a Word-based Concurrent Perturbation (WBCP) to automatically mine the attribution score for each word and use this as a measure of its degree of importance. Then we use the measure mentioned above to compute a word-level importance distribution as supervision information. Finally, we describe how to use the supervision information to refine the attention mechanism of the target model, improving the accuracy and robustness of text classification tasks.

\subsection{Word-based Concurrent Perturbation}
\label{sec:WBCP}
The basic assumption of our design is based on the following fact: under the premise of trying not to change the semantics of the input text, unimportant words can withstand more changes than more significant ones. Specifically, a little noise on keywords can lead to dramatic changes in the final results, while greater noise on the unimportant ones won't easily lead to changes in results. Therefore, we can estimate the importance distribution of the words according to the maximum amount of noise they can tolerate. To be specific, we try to concurrently add as much noise as possible to perturb each word embedding without changing the latent representations (e.g., the hidden states for classification) of the text and the prediction result. The above process can be optimized according to the maximum entropy principle.

Given a sentence consisting of $n$ words $s=\{w_1, w_2, ..., w_m\}$, we map each word into its embedding vector $\boldsymbol{X}=\{\boldsymbol{x}_1, \boldsymbol{x}_2, ..., \boldsymbol{x}_n\}$. Actually, WBCP (Word-based Concurrent Perturbation) is based on the embedding of each token $\boldsymbol{X}$ but not each word $s$. Intuitively, one word can be tokenized into several parts, and various parts have various influences on the representation. Considering that, in experiments, the perturbation is added to each token generated by the tokenizer, which means each token has its own $\sigma_{i}$ (maximum safety radius). For ease of explanation and comprehension, here we take the traditional embedding where $m=n$ (each word has only one embedding, e.g. word2vec, glove, and so on) as an example in Figure~\ref{Fig:perturb} and Section~\ref{sec:WBCP}. We assume that the noise on word embeddings obeys a Gaussian distribution $\boldsymbol{\epsilon}_{i} \sim \mathcal{N}\left(\mathbf{0}, \mathbf{\Sigma}_{i}=\sigma_{i}^{2} \mathbf{I}\right)$ and let $ \widetilde{\boldsymbol{x}_i}=\boldsymbol{x}_i+\boldsymbol{\epsilon}_{i} $ denote an input with noise $\boldsymbol{\epsilon}_{i}$. We use $\boldsymbol{h}$, $\boldsymbol{y}$ and $\widetilde{\boldsymbol{h}}$, $\widetilde{\boldsymbol{y}}$ to indicate the hidden state for classification and the prediction result of a pre-trained model with no noise and with noise respectively. Then we can write the loss function of WBCP as follows:
\begin{gather}
\begin{split}
\mathcal{L}_{WBCP}=||\widetilde{\boldsymbol{h}}-\boldsymbol{h}||^{2}_2+||\widetilde{\boldsymbol{y}}-\boldsymbol{y}||^{2}_2\\-{\lambda}{\sum}_{i=1}^n H({\boldsymbol{\epsilon}_{i}})|_{\boldsymbol{\epsilon}_{i} \sim \mathcal{N}\left(\mathbf{0}, \mathbf{\Sigma}_{i}=\sigma_{i}^{2} \mathbf{I}\right)},
\label{eq-perturb}
\end{split}
\end{gather}
where $ \lambda $ is a hyperparameter that balances the strength of noise.

The first and the second term of Eq.~(\ref{eq-perturb}) mean that we need to minimize the L2-normalized euclidean distance between the two hidden states and between the two predictions respectively, to quantify the change of information~\cite{jain2019attention}. The first term maintains latent representations to prevent modification of the text semantics, and the second term prevents excessive perturbations from causing the model to mispredict. The last term indicates that we need to maximize the entropy $ H(\boldsymbol{\epsilon}_{i})|_{\boldsymbol{\epsilon}_{i} \sim \mathcal{N}\left(\mathbf{0}, \mathbf{\Sigma}_{i}=\sigma_{i}^{2} \mathbf{I}\right)} $ to encourage adding as much noise as possible to each word embedding. We can simplify the maximum entropy of the Gaussian distribution as follows:
\begin{equation}
\begin{aligned}
& Maximize(H({\boldsymbol{\epsilon}}_{i})) \\
&= Maximize(-\int p({\boldsymbol{\epsilon}}_{i}) \ln p({\boldsymbol{\epsilon}}_{i}) d {\boldsymbol{\epsilon}}_{i})\\
&= Maximize(\frac{1}{2}(\ln(2{\pi}{\sigma_i}^{2})+1))\\
&= Maximize(\ln 2 (\frac{1}{2} \log (2 \pi e)+ \log {\sigma}_{i}))\\
&= Maximize(\log {\sigma}_{i})\nonumber
\label{eq-simplify}
\end{aligned}
\end{equation}
Finally we can use Eq.~(\ref{eq-simplify}) to rewrite our final objective function:
\begin{equation}
\begin{split}
\mathcal{L}_{WBCP}=||\widetilde{\boldsymbol{h}}-\boldsymbol{h}||^{2}_2+||\widetilde{\boldsymbol{y}}-\boldsymbol{y}||^{2}_2+{\lambda}{\sum}_{i=1}^n \log({-\sigma_i})
\label{eq-rewrite}
\end{split}
\end{equation}
The illustration of WBCP is given in Figure~\ref{Fig:perturb}. After fixing the parameters of the pre-trained model, the only learnable parameters ${\sigma}=\{ {\sigma}_1, {\sigma}_2, {\sigma}_3, {\sigma}_4 \}$ can be considered as the perturbation radii, which is positively associated with the perturbation amplitude. Specifically, the larger $ {\sigma}_i $ WBCP gets, the more likely $ {\boldsymbol{\epsilon}}_i $ is a big number, the more noise is added to $ \boldsymbol{x}_i $, and the less important it is. As what is shown in the picture, it is obvious that $ {\sigma}_2 > {\sigma}_1 > {\sigma}_4 > {\sigma}_3 $. According to the analysis listed above, we know that $ \boldsymbol{w}_2 $ (\textit{a}) is the least important word and $ \boldsymbol{w}_3 $ (\textit{nice}) is the most significant one, for $ \boldsymbol{x}_2 $ can tolerate the most noise while $ \boldsymbol{x}_3 $ can hardly stand any perturbation.

During the training stage of WBCP, $\sigma$ is first initialized as the normal distribution and then normalized by the standard deviation of sentence embeddings before generating noise. And we set the epochs to $500$ for most datasets. Actually, most perturbation models converge within less than $200$ steps, but we choose more epochs for the time cost is acceptable. However, IMDB's settings differ because of the large training and testing set. Therefore, we set epochs to 300 for it. As for the optimizer, we select AdamW with a learning rate of 0.01.

\subsection{Attention supervision}

We obtain the $\sigma$s, the perturbation magnitudes, by optimizing Eq.~(\ref{eq-rewrite}) on the pre-trained model. If a word embedding $\boldsymbol{x}_i$ can tolerate more noise without impacting the semantics of input text, $\sigma_i$ will be larger, which means the word $\boldsymbol{x}_i$ is less important. Conversely, small $\sigma_i$ indicates that slight perturbations of word embedding $\boldsymbol{x}_i$ will lead to semantic drift and may affect the classification result. We can therefore use the perturbation magnitude to compute a word-level importance distribution as attention supervision information, as shown below:
\begin{equation}
\begin{aligned}
\alpha'_i &= 1-\frac{\sigma_i}{\text{max}_{j}\{\sigma_j\}} \\
{\boldsymbol{\widetilde{\alpha}}} &= \text{Softmax}({\boldsymbol{\alpha'}})
\end{aligned}
\label{eq-importance}
\end{equation}
It is worth noting that our method generates sample-specific attention supervision, where the weight of each word is quantified according to the perturbation magnitude, instead of using the same importance weight for all words~\cite{tang2019progressive,choi2020lessismore}. Also, the quantification occurs in the embedding space rather than replacing the token with a predefined value, thus avoiding the OOD problem caused by masking schemes.

\begin{algorithm}[h!]
\caption{\textbf{Perturbation-based self-supervised attention}}
\KwIn{training dataset $D$, attention-based model $f(\cdot,\theta)$, the number of iterations $T$.}
Pre-train model $f(\cdot,\theta)$ on $D$ and update $\theta$ using Adam. \\
\For{$t = 1,...T$}{
    Fix $\theta$, and minimize WBCP objective function by Eq.~(\ref{eq-rewrite}) using Adam. \\
    Obtain the perturbation amplitude $\sigma$ for each sample in $D$. \\
    Calculate the attention supervision $\widetilde{\alpha}$ by Eq.~(\ref{eq-importance}) for each sample in $D$. \\
    Re-train model on $D$ with the attention supervision $\widetilde{\alpha}$ by Eq.~(\ref{eq-trainloss}) and update $\theta$ using Adam. \\
}
\label{flow}
\end{algorithm}

\subsection{Perturbation-based Self-supervised Attention}
\label{sec:perturbation_attention}
We do not use $\boldsymbol{\widetilde{\alpha}}$ to  generate a new attention distribution to replace the original one $\boldsymbol{{\alpha}}$. Rather, we use it as a supervision target for the attention weights. We want the attention supervision to make the model notice more words that have an influence on the output. In this way, some low-frequency context words with great importance that would normally be ignored can be discovered by attention learning. In this section, we describe how to exploit the supervision information $\widetilde{\alpha}$ to guide the learning of model attention strengths.

\begin{table*}[h!]
    \centering
    \caption{Detailed dataset statistics.}
    \resizebox{0.8\linewidth}{!}{
    \begin{tabular}{c|c|cccc}
    \toprule
    \textbf{Task} & \textbf{Dataset} & \textbf{Class} &  \textbf{AvgLen} & \textbf{Train} & \textbf{Test} \\
    \hline
    \multirow{6}{*}{Sentence Classification} & SST2~\cite{socher2013recursive} & 2 & 19 & 6,920 & 1821 \\
    & TREC~\cite{li2002learning} & 6 & 10 & 5,452 & 500 \\ 
    & MR~\cite{pang2005seeing} & 2 & 19 & 10,662 & -- \\
    & CR~\cite{hu2004mining} & 2 & 19 & 3,775 & -- \\
    & SUBJ~\cite{pang2004sentimental} & 2 & 23 & 10,000 & -- \\
    & MPQA~\cite{wiebe2005annotating} & 2 & 3 & 10,606 & -- \\
    \hline
    Document Categorization & IMDB~\cite{maas2011learning} & 2 & 280 & 25,000 & 25,000 \\
    \hline
    \multirow{3}{*}{Aspect-based Sentiment Analyis} & REST~\cite{pontiki2014semeval} & 3 & 16 & 3,591 & 1,121 \\
    & LAPTOP~\cite{pontiki2014semeval} & 3 & 17 & 2,292 & 639 \\
    & TWITTER~\cite{dong2014adaptive} & 3 & 19 & 6,248 & 692 \\
     \bottomrule
    \end{tabular}
    }
    \label{tab:data}
\end{table*}

\begin{table*}[h!]
    \centering
    \caption{Setup for Att-BiLSTM and Memory Net} 
    \resizebox{1.0\linewidth}{!}{
    \begin{tabular}{c|c|cc}
    \toprule
    \textbf{Task} & \textbf{Dataset} & \textbf{Dimension of hidden states} &  \textbf{Dimension of attention context} \\
    \hline
    \multirow{6}{*}{Sentence Classification} & SST2~\cite{socher2013recursive} & 150 & 100 \\
    & TREC~\cite{li2002learning} & 150 & 50 \\ 
    & MR~\cite{pang2005seeing} & 150 & 100 \\
    & CR~\cite{hu2004mining} & 150 & 50 \\
    & SUBJ~\cite{pang2004sentimental} & 150 & 100 \\
    & MPQA~\cite{wiebe2005annotating} & 150 & 100 \\
    \hline
    Document Categorization & IMDB~\cite{maas2011learning} &150 & 300 \\
    \hline
    \multirow{3}{*}{Aspect-based Sentiment Analyis} & REST~\cite{pontiki2014semeval} & 300 & 300 \\
    & LAPTOP~\cite{pontiki2014semeval} & 300 & 300 \\
    & TWITTER~\cite{dong2014adaptive} & 300 & 300 \\
     \bottomrule
    \end{tabular}
    }
    \label{tab:hyper}
\end{table*}

Our method is shown in Algorithm~\ref{flow}. We first pre-train an attention-based model $f(\cdot,\theta)$ based on the classification dataset $D$. We then fix the model parameters $\theta$ and minimize the WBCP objective using Eq.~(\ref{eq-rewrite}) to obtain the perturbation amplitude $\sigma$ for each sample, and used to compute the attention supervision $\widetilde{\alpha}$ using Eq.~(\ref{eq-importance}). We then retrain the model using $\widetilde{\alpha}$ to guide the attention distribution $\alpha$ produced by the model. The above process can iterate $T$ times to capture the important distribution more accurately. The training objective function with attention supervision $\widetilde{\alpha}$ is defined as follows:
\begin{gather}
\begin{split}
\mathcal{L}_{cls} = \frac{1}{M}{\sum}^{M}_{m=1}\hat{y}_m\log y_m+\gamma \text{KL}(\widetilde{\alpha}_m || \alpha_m),
\label{eq-trainloss}
\end{split}
\end{gather}
where $M$ is the number of samples, $ \gamma $ is a hyperparameter that controls the strength of attention supervision, $\hat{y}_m$ and $y_m$ are the ground-truth label and predicted output for the $m$-th sample respectively. The first term is the Cross-Entropy Loss for classification, and the second term is the Kullback–Leibler Divergence between the distributions of attention $ \alpha_m$ produced by model and attention supervision information $ \widetilde{\alpha}_m $ for the $m^{th}$ sample.

It's worth noting that our method requires extra computations, but the time cost is usually acceptable because nearly all the process is parallel. The analysis are explained in Appendix~\ref{sec:extra-com}.

\section{Experiments}

We tried PBSA on several text classification tasks, including sentence classification, document categorization, and aspect-level sentiment analysis. Experimental results demonstrate that PBSA consistently enhances the performance and robustness of various attention-based baselines, and outperforms some strong models following self-supervised attention learning. Furthermore, a visualization analysis confirms that our model is capable of generating high-quality attention for target tasks. We aim to answer the following questions:
\begin{list}{}{}
\item{\textbf{RQ1:} Does PBSA improve model accuracy?}
\item{\textbf{RQ2:} Is PBSA more effective than other approaches?}
\item{\textbf{RQ3:} How do hyperparameters affect the results?}
\item{\textbf{RQ4:} How does PBSA work?}
\end{list}

\subsection{Datasets and Baselines}\label{dataset}

The statistics of widely-studied datasets used by different tasks are listed in Table~\ref{tab:data}. These datasets come from different topics, such as movie reviews, customer reviews, social reviews, and question type. In particular, since there is no standard partition of MR, CR, SUBJ, and MPQA, we follow the data splitting protocol, 7:1:2 for them to get the training, validation, and test sets. For the aspect-level tasks, we remove the instances with conflict sentiment labels in Laptop and Restaurant as implemented in~\cite{2017Recurrent}.

As for hyperparameters, we use a grid search to find the optimal value of $\gamma$ and $T$ for each dataset, from the sets $\gamma\in\{0.05,0.1,1.0,2.0,10,100\}$ and $T\in\{1,2,3,4\}$. We use the Adam optimizer with learning rate 0.001 and the batch size is set to 64.

We use Att-BiLSTM, Memory Network, BERT, DEBERTA, ELECTRA, Att-BERT, BERTABSA, Att-BERTABSA as baselines and explain the details about them in Appendix~\ref{sec:baselines}.

The setup of hyperparameters for Att-BiLSTM and Memory Net are listed in Table~\ref{tab:hyper}. To make a fair compare with other algorithms, we set our hyperparameters the same as theirs.

\begin{table*}[t!]
\centering
\caption{The performance of PBSA on the document-level and sentence-level classification.}
\resizebox{0.9\linewidth}{!}{
\begin{tabular}{@{}c|ccccccc|c@{}}
\toprule
Model            & IMDB           & SST2           & TREC          & MR             & CR             & SUBJ           & MPQA           & Average        \\ \midrule
Att-BiLSTM          & 87.21          & 83.42          & 90.60          & 77.04          & 76.82          & 89.82          & 70.59          & 82.20          \\
Att-BiLSTM+PBSA & \textbf{89.14} & \textbf{85.72} & \textbf{92.20} & \textbf{79.05} & \textbf{77.64} & \textbf{90.53} & \textbf{71.31} & \textbf{83.65} \\
\hline
Att-BERT(base) & 92.53 & 91.43 & 96.60 & 79.26 & 89.06 & 94.30 & 89.69 & 90.41 \\
Att-BERT(base)+PBSA & \textbf{92.61} & \textbf{91.93} & \textbf{97.20} & \textbf{79.97} & \textbf{89.38} & \textbf{94.76} & \textbf{90.21} & \textbf{90.86} \\
\hline
BERT(base)          & 92.92          & 91.71          & 96.60          & 85.47          & 89.42          & 96.30          & 89.59         & 91.71         \\
BERT(base)+PBSA & \textbf{93.48} & \textbf{92.20} & \textbf{97.80} & \textbf{86.08} & \textbf{90.21} & \textbf{97.50} & \textbf{90.57} & \textbf{92.55}\\
\hline
DEBERTA(base) & 91.14 & 92.69 & 96.20	& 86.64	& 91.01	& 95.30	& 89.74 & 91.82\\
DEBERTA(base)+PBSA & \textbf{91.68} & \textbf{93.02}	& \textbf{96.80}	& \textbf{87.18}	& \textbf{91.80}	& \textbf{95.70}	& \textbf{90.82} & \textbf{92.43}\\
\hline
ELECTRA(base) & 93.48 & 94.67 & 96.8 & 89.27 & 92.2	& 96.75 & 90.9 & 93.44\\
ELECTRA(base)+PBSA & \textbf{93.87} & \textbf{95.43}	& \textbf{97.40}	& \textbf{89.88}	& \textbf{92.99}	& \textbf{97.30}	& \textbf{91.55} & \textbf{94.06}\\

\bottomrule
\end{tabular}
}
\label{tab:doc-cls}
\end{table*}

\begin{table*}[t!]
\centering
\caption{Experimental accuracy on the document-level and sentence-level classification compared with others}
\resizebox{0.9\linewidth}{!}{
\begin{tabular}{@{}c|ccccccc|c@{}}
\toprule
Model            & IMDB           & SST2           & TREC          & MR             & CR             & SUBJ           & MPQA           & Average        \\ \midrule
Att-BiLSTM          & 87.21          & 83.42          & 90.60          & 77.04          & 76.82          & 89.82          & 70.59          & 82.20          \\
+Gradient & 86.79 & 85.06 & 91.20 & 77.60 & 76.54 & 89.82 & 70.76 & 82.53 \\
+SANA~\cite{choi2020lessismore} & 88.03 & 84.35 & - & - & - & - & - & - \\
+PBSA & \textbf{89.14} & \textbf{85.72} & \textbf{92.20} & \textbf{79.05} & \textbf{77.64} & \textbf{90.53} & \textbf{71.31} & \textbf{83.65} \\
\bottomrule
\end{tabular}
}
\label{tab:doc-com}
\end{table*}

\begin{table*}[t!]
\centering
\caption{The performance of PBSA on the aspect-level classification.}
\resizebox{0.9\linewidth}{!}{
\begin{tabular}{l|c|ccccc}
\toprule
Models & \multicolumn{2}{c|}{REST}        & \multicolumn{2}{c|}{LAPTOP}                                             & \multicolumn{2}{c}{TWITTER}                        \\ \cline{2-7} 
\multicolumn{1}{c|}{}      & Accuracy      & \multicolumn{1}{c|}{Macro-F1}      & \multicolumn{1}{c|}{Accuracy}      & \multicolumn{1}{c|}{Macro-F1}      & \multicolumn{1}{c|}{Accuracy}      & Macro-F1      \\ \hline
 MN~\cite{wang2018target}   & 77.32 & \multicolumn{1}{c|}{65.88}   & \multicolumn{1}{c|}{68.90}   & \multicolumn{1}{c|}{63.28}         & \multicolumn{1}{c|}{67.78}         & 66.18         \\
MN (Ours)              & 79.89         & \multicolumn{1}{c|}{65.89}         & \multicolumn{1}{c|}{72.68}         & \multicolumn{1}{c|}{61.97}         & \multicolumn{1}{c|}{68.34}         & 66.23         \\
+PBSA        & \textbf{83.98}         & \multicolumn{1}{c|}{\textbf{70.84}}         & \multicolumn{1}{c|}{\textbf{75.75}}         & \multicolumn{1}{c|}{\textbf{67.21}}         & \multicolumn{1}{c|}{\textbf{72.10}}         & \textbf{69.64}         \\

\hline
BERTABSA       & 79.80        & \multicolumn{1}{c|}{71.37}         & \multicolumn{1}{c|}{79.38}         & \multicolumn{1}{c|}{75.69}         & \multicolumn{1}{c|}{76.01}         & 74.52        \\
+PBSA        & \textbf{79.89}         & \multicolumn{1}{c|}{\textbf{71.59}}         & \multicolumn{1}{c|}{\textbf{79.51}}         & \multicolumn{1}{c|}{\textbf{75.87}}         & \multicolumn{1}{c|}{\textbf{76.11}}         & \textbf{74.69}         \\
\hline

Att-BERTABSA       & 83.29        & \multicolumn{1}{c|}{75.87}         & \multicolumn{1}{c|}{77.98}         & \multicolumn{1}{c|}{75.02}         & \multicolumn{1}{c|}{73.99}         & 71.23        \\
+PBSA        & \textbf{83.41}         & \multicolumn{1}{c|}{\textbf{76.70}}         & \multicolumn{1}{c|}{\textbf{78.65}}         & \multicolumn{1}{c|}{\textbf{75.53}}         & \multicolumn{1}{c|}{\textbf{74.45}}         & \textbf{72.88}         \\
\bottomrule
\end{tabular}
}
\label{tab:aspect-cls}
\end{table*}

\begin{table*}[t!]
\centering
\caption{Experimental results on aspect-level tasks compared with others.}
\resizebox{0.9\linewidth}{!}{
\begin{tabular}{l|c|ccccc}
\toprule
Models & \multicolumn{2}{c|}{REST}        & \multicolumn{2}{c|}{LAPTOP}                                             & \multicolumn{2}{c}{TWITTER}                        \\ \cline{2-7} 
\multicolumn{1}{c|}{}      & Accuracy      & \multicolumn{1}{c|}{Macro-F1}      & \multicolumn{1}{c|}{Accuracy}      & \multicolumn{1}{c|}{Macro-F1}      & \multicolumn{1}{c|}{Accuracy}      & Macro-F1      \\ \hline
 MN~\cite{wang2018target}   & 77.32 & \multicolumn{1}{c|}{65.88}   & \multicolumn{1}{c|}{68.90}   & \multicolumn{1}{c|}{63.28}         & \multicolumn{1}{c|}{67.78}         & 66.18         \\
MN (Ours)              & 79.89         & \multicolumn{1}{c|}{65.89}         & \multicolumn{1}{c|}{72.68}         & \multicolumn{1}{c|}{61.97}         & \multicolumn{1}{c|}{68.34}         & 66.23         \\
+Gradient~\cite{serrano2019attention}       & 76.85         & \multicolumn{1}{c|}{60.06}         & \multicolumn{1}{c|}{71.11}         & \multicolumn{1}{c|}{63.53}         & \multicolumn{1}{c|}{67.77}         & 64.91         \\
+AWAS~\cite{tang2019progressive}         & 78.75         & \multicolumn{1}{c|}{69.15}         & \multicolumn{1}{c|}{70.53}         & \multicolumn{1}{c|}{65.24}         & \multicolumn{1}{c|}{69.64}         & 67.88         \\
+Boosting~\cite{su2021enhanced}             & 77.66         & \multicolumn{1}{c|}{66.23}         & \multicolumn{1}{c|}{69.28}         & \multicolumn{1}{c|}{64.17}         & \multicolumn{1}{c|}{68.14}         & 67.12         \\
+Adaboost~\cite{su2021enhanced}             & 76.77         & \multicolumn{1}{c|}{62.29}         & \multicolumn{1}{c|}{67.88}         & \multicolumn{1}{c|}{60.52}         & \multicolumn{1}{c|}{66.96}         & 65.09         \\
+PGAS~\cite{su2021enhanced} & 78.98         & \multicolumn{1}{c|}{69.42}         & \multicolumn{1}{c|}{70.84}         & \multicolumn{1}{c|}{65.58}         & \multicolumn{1}{c|}{69.78}         & 67.80         \\
\hline
+PBSA        & \textbf{83.98}         & \multicolumn{1}{c|}{\textbf{70.84}}         & \multicolumn{1}{c|}{\textbf{75.75}}         & \multicolumn{1}{c|}{\textbf{67.21}}         & \multicolumn{1}{c|}{\textbf{72.10}}         & \textbf{69.64}         \\
\bottomrule
\end{tabular}
}
\label{tab:aspect-com}
\end{table*}

\subsection{\textbf{RQ1:} Sentence-level and Document-level Classification}\label{sec:sc}
\begin{figure*}[t!]
	\centering
	\includegraphics[width=1.0\linewidth]{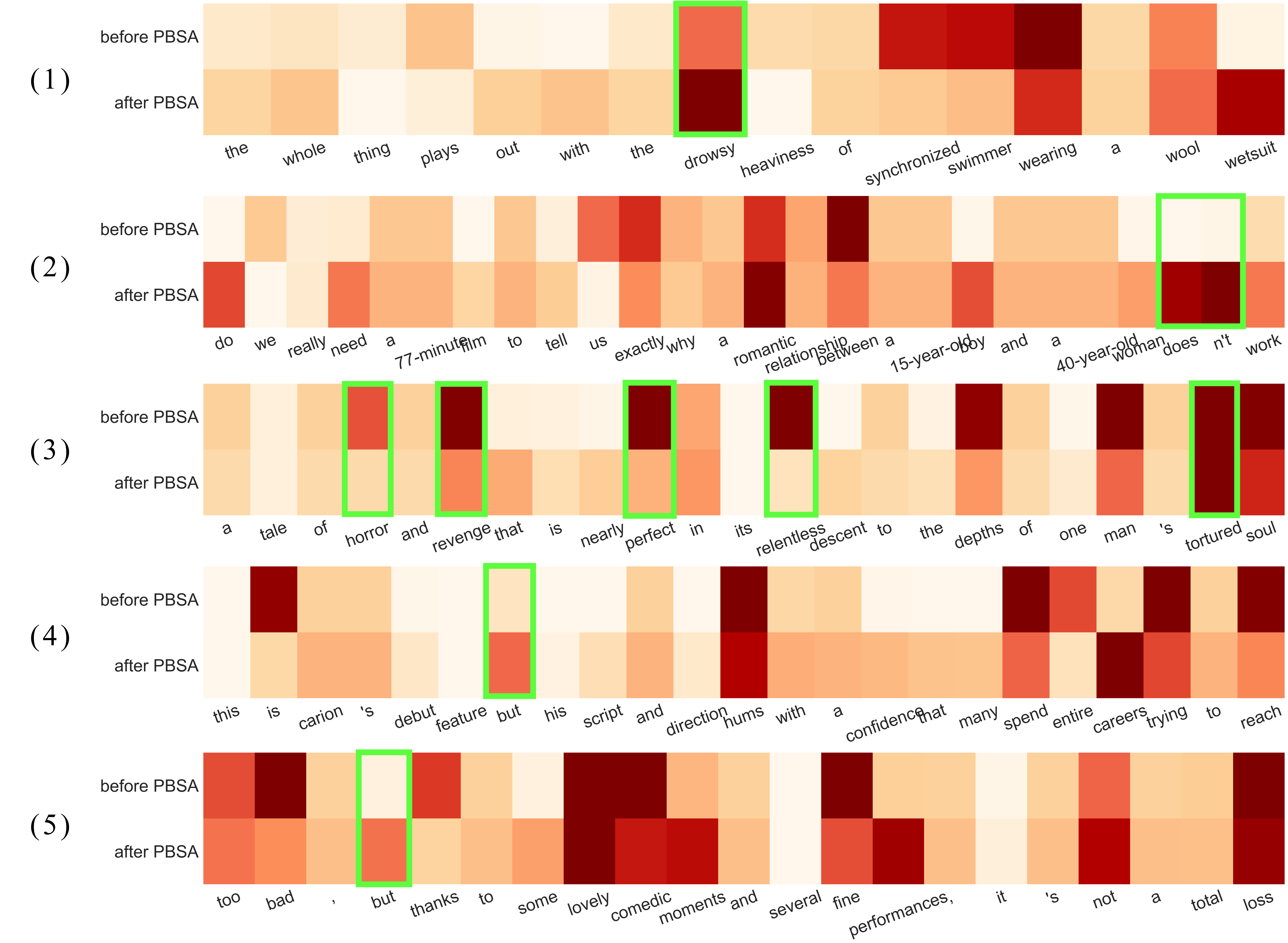}
	\caption{The visualization result of several samples on SST2 test set.}
    \label{Fig:visual}
\end{figure*}

To verify that PBSA can improve the performance of the attention-based model, in this section, we use the classic Att-BiLSTM~\cite{bilstm2016zhou} and the pre-trained models BERT~\cite{devlin2018bert}, DEBERTA~\cite{he2020deberta}, and ELECTRA~\cite{clark2020electra} as the baselines. It is worth noting that Transformers use multiple-layer and multiple-head attention, so selecting the suitable head as the supervised target is difficult~\cite{su2021enhanced}. Hence, how to effectively combine its multiple-layer and multiple-head attention with our method is an exciting and valuable question.

The previous researchers have yet to find a good way to apply their methods to Transformers, and we have made some explorations in this field, which is also one of our innovations. We explore two simple strategies to combine our approach with Transformers, 1) We first add a scaled dot-product attention layer to the output of BERT to derive a fixed-sized sentence representation for classification, and we call this model Att-BERT for short. 2) We also try a simple but effective way to combine the internal multi-head attention in Transformers with our method. Specifically, we average the multi-head attention of all the layers and compress the attention matrix to a vector to be guided by our mechanism.

Table~\ref{tab:doc-cls} reports the experimental results on the seven datasets of sentence classification and document categorization. We observe that our method consistently helps improve the accuracy of the baseline on all the datasets. The average accuracy of our approach on the five baselines across seven datasets are 83.65, 90.86, 92.55, 92.43, and 94.06, an improvement of 1.44\%, 0.45\%, 0.83\%, 0.66\%, and 0.66\% over the baselines (82.21, 90.41, 91.71, 91.82, and 93.44). The results demonstrate that our approach delivers significant performance improvements over the baselines. It also indicates that the current model limits the potential of attention mechanisms when without any supervision information. However, PBSA can mine the potential important words and then guide the attention mechanism of the model to learn a good inductive bias.

However, we find the improvements on pre-trained models are relatively marginal compared with smaller models like Att-BiLSTM. The phenomenon indicates that the pre-training on large corpora relieves the attention bias to some extent, which is further verified in Section~\ref{sec:ratio}. Moreover, we also find the size of the pre-trained model also impacts the performance of PBSA. We conduct the experiments on BERT-small and ELECTRA-small (shown in Table~\ref{tab:small-cls}), and PBSA gains greater improvements under the same settings. To sum up, the attention bias may be more likely to appear in smaller-size models and smaller-scaled datasets. And the performance of PBSA will be more significant in these scenarios.

\begin{table*}[t!]
\centering
\caption{The performance of PBSA on small-size pre-trained models.}
\resizebox{0.9\linewidth}{!}{
\begin{tabular}{@{}c|ccccccc|c@{}}
\toprule
Model            & IMDB           & SST2           & TREC          & MR             & CR             & SUBJ           & MPQA           & Average        \\ \midrule
BERT(small)          & 90.81          & 88.85          & 95.60          & 81.16          & 81.61          & 94.45          & 87.23         & 88.53         \\
BERT(small)+PBSA & \textbf{91.73} & \textbf{90.17} & \textbf{97.40} & \textbf{82.33} & \textbf{83.07} & \textbf{96.20} & \textbf{88.54} & \textbf{89.92}\\
\hline
ELECTRA(small) & 92.37 & 91.21 & 96.00 & 83.60 & 87.30	& 95.70 & 88.97 & 90.74\\
ELECTRA(small)+PBSA & \textbf{93.35} & \textbf{92.20}	& \textbf{96.40}	& \textbf{84.72}	& \textbf{88.62}	& \textbf{96.65}	& \textbf{90.62} & \textbf{91.79}\\

\bottomrule
\end{tabular}
}
\label{tab:small-cls}
\end{table*}

\begin{figure*}[h!]
	\centering
	\includegraphics[width=0.9\linewidth]{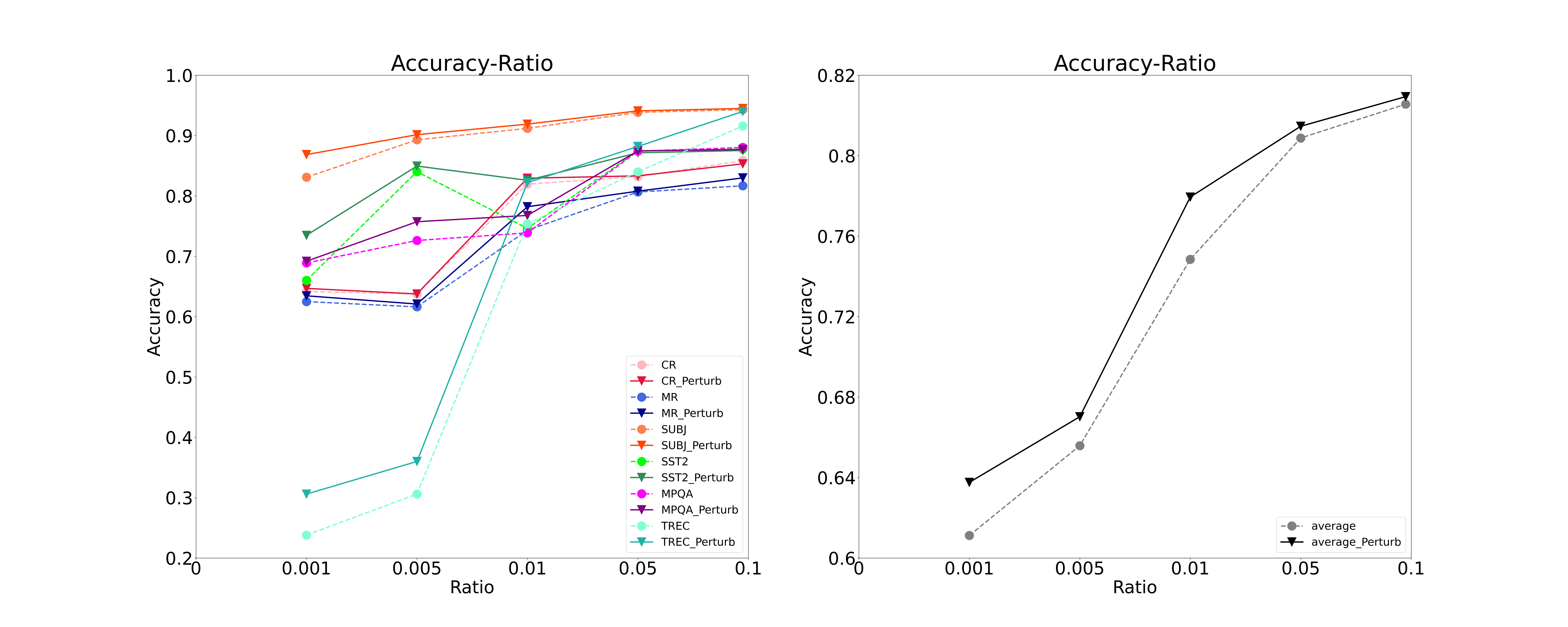}
    \caption{The chart of the fluctuations of accuracy when we change the value of the sample ratio. Each triangle point and circular point corresponds to the accuracy of BERT and BERT+PBSA under the current sample ratio, respectively.}
    \label{Fig:ratio}
\end{figure*}

\subsection{\textbf{RQ1:} Aspect-level Sentiment Analyis}

To further verify the effectiveness of our approach, we apply PBSA into MN~\cite{mn2016tang,wang2018target}, BERTABSA~\cite{DBLP}, and Att-BERTABSA~\cite{su2021enhanced}. Both BERTABSA and Att-BERTABSA are typical and simple ways to apply BERT to aspect-level classification tasks. The difference is that BERTABSA directly uses the hidden states of the aspect words to classify, while Att-BERTABSA adds an attention layer to the output of BERT. To show that our method truly improves the results, we only use the most critical parts of the model without any other tricks or mechanisms (e.g. the gating mechanism). We conduct experiments on three benchmark datasets of aspect-based sentiment analysis and  PBSA outperforms all the baselines on all datasets both in accuracy and Macro-F1. As shown in Table~\ref{tab:aspect-cls}, compared with other tasks, PBSA has a more significant improvement on these small-scale datasets, indicating that the original attention lacks a good inductive bias due to limited labeled data. With the help of PBSA, the robustness of the model can be improved effectively.

\subsection{\textbf{RQ1:} Performances under Different Sample Ratios}
\label{sec:ratio}
To verify the performance of our approach on low-resource tasks, we conduct experiments on different values of sample ratio. We get sample sets from the original datasets with sample ratio $\in\{0.001, 0.005, 0.01, 0.05, 0.1\}$, and measure the performances of BERT and BERT+PBSA on these sample sets according to their accuracy.

As shown in Figure~\ref{Fig:ratio}, the performances of BERT and BERT+PBSA have the same trend. As the accuracy of BERT increases, the accuracy of BERT+PBSA increases and vice versa. As explained in Section~\ref{sec:perturbation_attention}, the attention supervision information is obtained based on the pre-trained model, whose performance has a direct influence on the quality of the attention supervision information and further affects the results of re-training. That may explain the strong correlation between the performance of BERT and BERT+PBSA.

The improvement is more prominent when the ratio is in the middle range (sample ratio$\in (0.005, 0.05)$). As listed above, when the ratio is small, the pre-trained model has a bad performance, which results in meaningless attention supervision information and further limits the performance of PBSA. As the value of the sample ratio increases, the original model performs better, and the quality of attention supervision information is enhanced, and then PBSA improves the model even more. However, the improvement is not without limitation. As the value of the sample ratio exceeds a certain value, the phenomenon of attention bias is no longer evident, and the improvement reduces. It may be because BERT is pre-trained on a large-scale corpus, and when we fine-tune it, its attention fits well on these 'larger-scale' sample sets, which makes the original model has scant room for improvement.

To sum up, the distribution of the attention parameters is not stable enough when the data is limited or the model size is small, which can be refined by PBSA. And the performance and lifting area of PBSA are closely related to the performance of the baseline.

\subsection{\textbf{RQ2:} Comparison with other methods}
On the tasks listed above, we compare our method with other advanced self-supervised attention learning approaches. SANA~\cite{choi2020lessismore} generates counterfactuals by a masking scheme and measures the difference in the softmax probability of the model between the counterfactual and original sample as an indicator of important words. AWAS~\cite{tang2019progressive} and PGAS~\cite{su2021enhanced} progressively mask the word with the largest attention weight or partial gradient. 
Most of these works don't publish their critical code and do their experiment only on certain specific tasks, so we directly compare our algorithm with their best results published on different tasks respectively. To make a fair comparison, we use the same word embedding and the same settings of hidden size to reproduce their baselines, which is listed in Table~\ref{tab:hyper}.

On the document-level and sentence-level tasks (Table~\ref{tab:doc-com}), PBSA is superior to SANA by 1.11\% and 1.37\%, which verifies that the word-based concurrent perturbation can mine the importance distribution of words more accurately than the masking scheme. On the aspect-level task (Table~\ref{tab:aspect-com}), compared with AWAS and PGAS, our method improves the model more. As we mentioned in the Introduction (Section~\ref{sec:intro}), our method can generate word-specific attention supervision while others treat the important words equally without discrimination. We speculate that this may be one of the main reasons for our improvement.

\begin{figure*}[h!]
	\centering
	\includegraphics[width=0.8\linewidth]{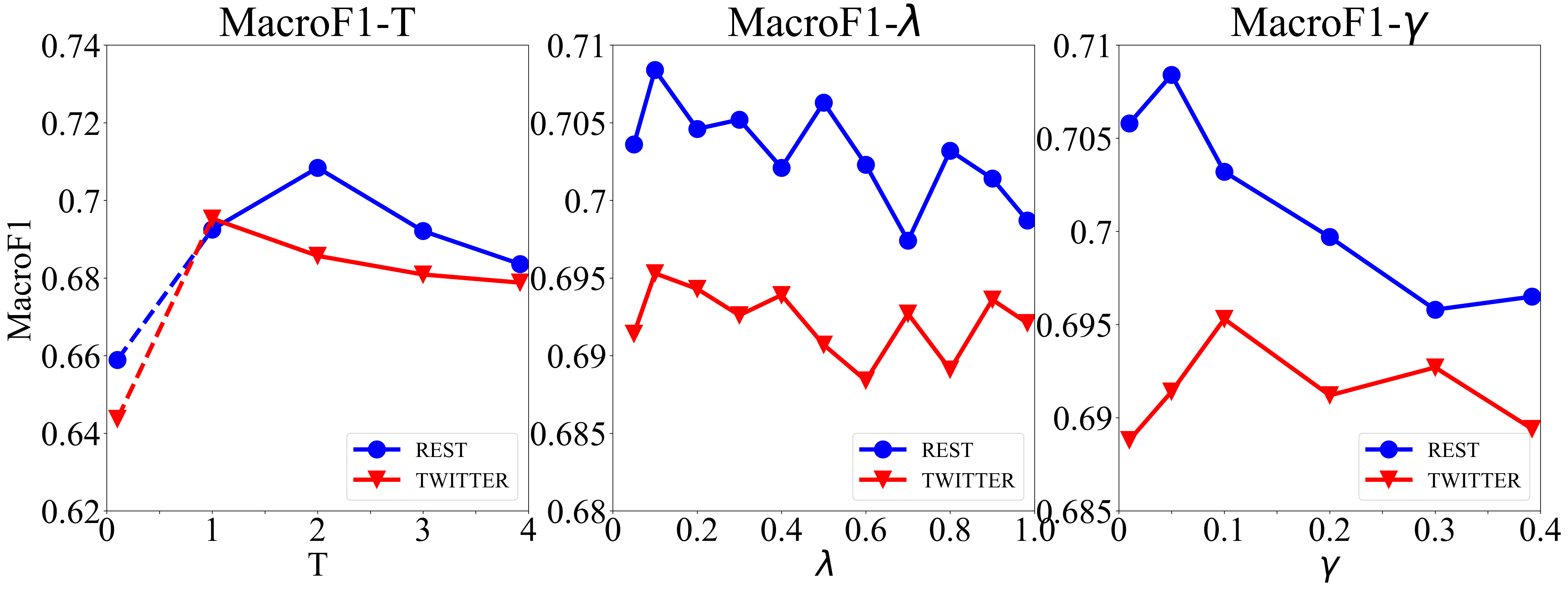}
    \caption{The chart of the fluctuations of Macro-F1 when we change the values of hyperparameters.}
    \label{Fig:hyper}
\end{figure*}

\subsection{\textbf{RQ2:} Comparison with human intuition methods}
From the aspect of human intuition, the gradient-based methods and leave-one-out methods are usually used to improve the interpretability of model. The current self-supervised attention learning methods are mostly based on word masking, which can be seen as a variation of leave-one-out methods. We also try to use the gradient-based method~\cite{serrano2019attention} to generate supervision information. As shown in Table~\ref{tab:doc-cls} and Table~\ref{tab:aspect-cls}, the gradient-based method does badly on most of the datasets, especially on aspect-level datasets. These results demonstrate that although the gradient-based method can improve the interpretability of the model, it does not necessarily improve the performance. However, our method enhances interpretability while also improving its performance.

\subsection{\textbf{RQ3:} Hyperparameter sensitivity}
As shown in Figure~\ref{Fig:hyper}, our method achieves the best results on REST and TWITTER when $T=2$ and $T=1$ respectively. When the increase of $T$, the performance increases initially and then decreases due to over-fitting. The performance of models won't change sharply with the increase of $T$ once they achieve the best results. In practice, we find that one iteration has achieved promising results.
The hyperparmeter $\lambda$ controls the perturbation degree of WBCP, when $\lambda$ is too large, it will deteriorate performance due to injecting too much noise. In all of our experiments, we set $\lambda$ as $0.1$. The hyperparmeter $\gamma$ controls the strength of attention supervision, when $\gamma$ is too large, it easily leads to overly penalize the alignment between the model attention and perturbation attention, which may hurt the model's internal reasoning process.

Compared with $\gamma$, $\lambda$ has less effect on results when the value of which changes slightly, but we cannot remove ${\sum}_{i=1}^n \log({-\sigma_i})$ from our loss function. Otherwise, the model will try not to add any noise to $x$ without the term, which makes PBSA get a meaningless supervision distribution that varies dramatically for the same sentence each time (the distribution is supposed to be essentially unchanged for the same sentence). On the other hand, results are more sensitive to $\gamma$, which determines if the models can reach the peak of the results.

\subsection{\textbf{RQ4:} Visualization analysis}
In this section, we select several attention visualizations on SST2 test set to explain how PBSA works. As shown in Figure~\ref{Fig:visual}, we see that PBSA \textbf{makes the model pay more attention to important but low-frequency words, reduces the focus on high-frequency words that do not affect the results, increases the difference in weight between words with conflicting meanings, and increases sensitivity to adversative relations in sentences.}

\paragraph{\textbf{Pay more attention to important but low-frequency words}}
Some words do have important effects on the results, but if they do not appear frequently enough then the traditional attention mechanism may not pay enough attention to them. As shown in Figure~\ref{Fig:visual}-(1), the word \textit{drowsy} has an important influence on the emotional polarity of the film. However, it is a low-frequency word in the corpus, which makes the attention mechanisms do not allocate enough weights to it, resulting in a classification error. After being trained by PBSA, the model can assign enough weights to \textit{drowsy}, which changes the result from false to correct.

\paragraph{\textbf{Reduce the focus on high-frequency words that do not affect the results}}
In baseline, some high-frequency words which do not contain any emotional polarity usually get high weights, while some important words that should have been focused on are ignored. As Figure~\ref{Fig:visual}-(2) shows, \textit{romantic} and \textit{doesn't} are words with strong emotional polarity. However, the baseline assigns greater weights to other high-frequency words (e.g., \textit{between}) with no emotional polarity, and thus ignores the words \textit{romantic} and \textit{doesn't} which results in misclassification. After being trained by PBSA, the model reduces the focus on \textit{between} and the weights allocated to the significant words increase correspondingly, which turns the result.

\paragraph{\textbf{Increase the difference in weight between words with conflicting meanings}}
As shown in Figure~\ref{Fig:visual}-(3), the baseline focuses on too many words: \textit{horror}, \textit{revenge}, \textit{perfect}, \textit{relentless}, \textit{torture}, and so on. Maybe all of the words are important but the meanings of them are conflicting, which interferes with the classification task. The model feels confused because it does not know how to make a prediction according to so many emotional words. After being trained by PBSA, the difference in the weight of emotional words becomes larger, which makes it get the right result. It should be noted that the entropy of attention distribution may not decrease because PBSA keeps attention to important words while diluting the distribution of the other words.

\paragraph{\textbf{Be more sensitive to adversative relations in sentences}}
If there are adversative conjunctions (e.g., but, however, and so on) in the sentence, it is likely to express two opposite emotions before and after the adversative conjunction. This is when the model needs to keenly feel the changes of emotional polarity in the sentence. From this aspect, the model is also supposed to assign higher weights to those adversative conjunctions. Judging from our results, it is unfortunate that the original attention mechanism tends to ignore these conjunctions for they seem to have no effect on results outwardly. As Figure~\ref{Fig:visual}-(4) and Figure~\ref{Fig:visual}-(5) show, the baseline ignores the word \textit{but} and results in errors. After being trained by PBSA, the baseline pays more attention to \textit{but} which makes both of the emotions before and after the adversative conjunction can be taken into consideration.

\section{Conclusions and future work}\label{Conclusion}
In this paper, we propose a novel self-supervised attention learning method based on word-based concurrent perturbation. The algorithm adds as much as noise to each word in the sentence under the premise of unchanged semantics to mine the supervision information to guide attention learning. Our experiments demonstrate that our method achieves significant performance improvements over the baselines on several text classification tasks. Moreover, we use several visualization samples to interpret how our method guides the internal reasoning process of models.

It is worth to note that we combine our method with transformers, which is not implemented in most of the previous attention guiding methods. Our strategies may not be the best ways to apply our algorithm into transformers, but they still prove the effectiveness of the proposed method. We will try to find more appropriate and effective strategies and incorporate our algorithm into other NLP tasks in the future.



\begin{appendices}

\section{Analysis of the Extra Computations}
\label{sec:extra-com}
The extra computations mainly come from the process of generating supervision information (Pre-training and re-training are the same as the common training method). The extra time required depends on the size of the model, the number of the samples and the epochs of training perturbation model. It is acceptable for most datasets because the whole process is parallel. All the sub-perturbation models have independent samples and training processes, and they just share the same pre-trained model whose parameters are fixed during the generating process. Therefore, the whole process can be handled concurrently if having enough GPU resources.

For SST2, TREC, MR, CR, SUBJ, and MPQA, the generating process (batch-size=64) can be finished on 2 * GTX 3090 within less than 15 min. Some small datasets (e.g. SST2, TREC and CR) only need 8 min to generate supervison information. However, as for IMDB, the number of samples is enormous, and their average length is too long. Therefore, we must use several GPUs (2 * GTX 3090 and 4 * GTX 1080ti) to simultaneously deal with each part of the dataset to finish the task in a limited time.

\section{Details of Baselines}
\label{sec:baselines}

The details of our baselines are listed below.

\begin{figure}[!h]
    \centering
    \includegraphics[width=1.0\linewidth]{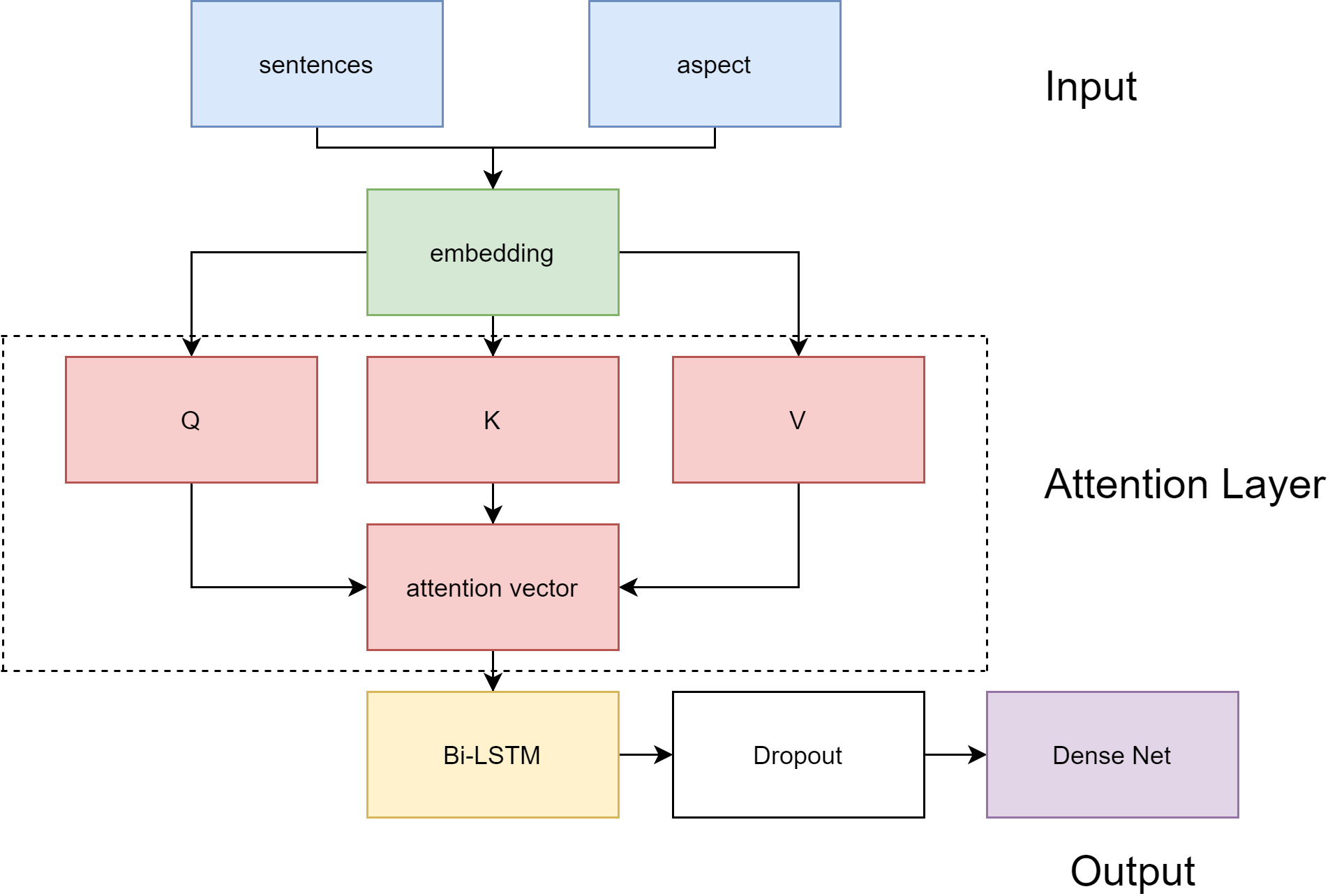}
    \caption{The illustration of Att-BiLSTM.}
    \label{Fig:Att-BiLSTM}
	\includegraphics[width=1.0\linewidth]{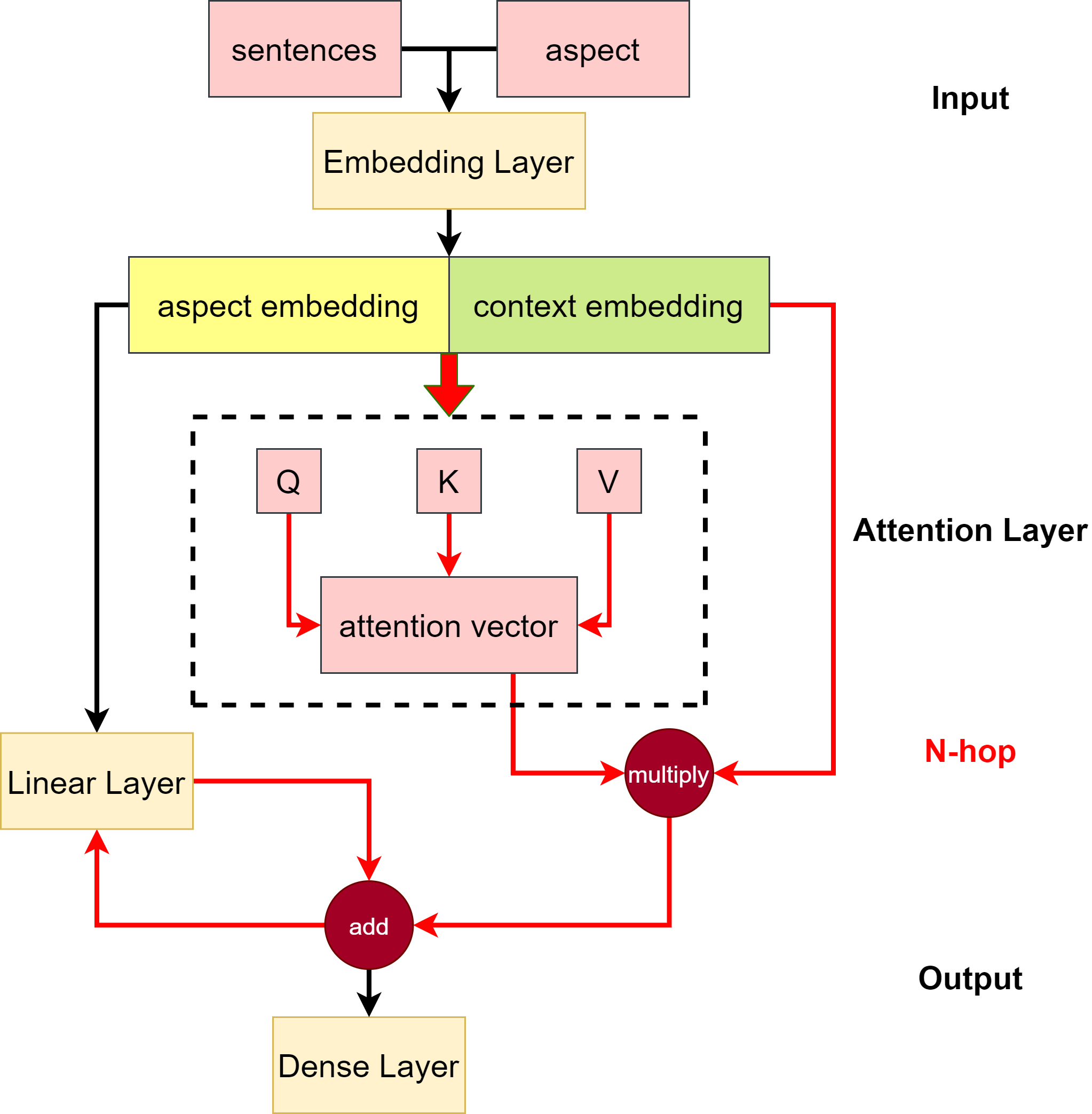}
    \caption{The illustration of Memory Net.}
    \label{Fig:MN}
\end{figure}

\subsection{Att-BiLSTM}
~\label{sec:Att-BiLSTM}
Figure~\ref{Fig:Att-BiLSTM} shows the structure of Att-BiLSTM. Att-BiLSTM first map each word into pre-trained skip-gram~\cite{2013Distributed} word embedding and then utilize 1-layered BiLSTM with a scale-dot attention mechanism to get sentence-level hidden states which are finally used for classification.

\subsection{Memory Network}
~\label{sec:Memory Network}
Figure~\ref{Fig:MN} shows the structure of MN. Memory Network uses an iteratively updated vector $A$ (initialized as the aspect embedding) and the context embedding to generate the attention distribution, which is then used to select the important information from the context embedding and iteratively update the vector $A$.

\subsection{Att-BERT}
~\label{sec:Att-BERT}
Figure~\ref{Fig:Att-BERT} shows the structure of Att-BERT. We add a scale-dot attention layer to the output of the BERT and use the output of the attention layer to classify.

\subsection{BERTABSA}
~\label{sec:BERTABSA}
Figure~\ref{Fig:BERTABSA} shows the structure of BERTABSA. We input the whole sentence to get the context representation of the aspect words, which is directly used for classification. To verify that our method truly improves the results, we delete the gating mechanism and use bert-base-uncased instead of bert-large-uncased.

\subsection{Att-BERTABSA}
~\label{sec:Att-BERTABSA}
\begin{figure}[!h]
    \centering
    \includegraphics[width=0.8\linewidth]{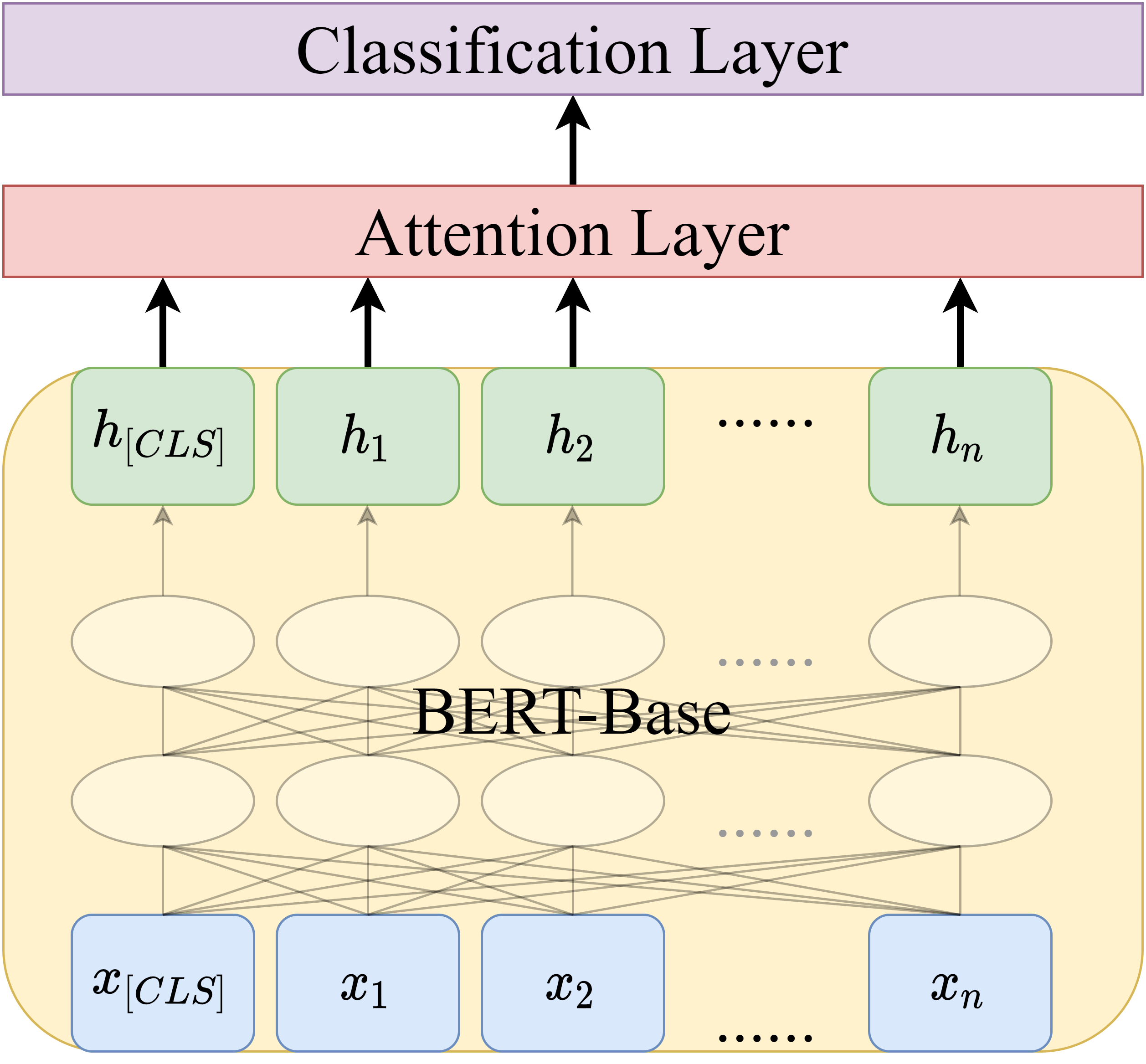}
    \caption{The illustration of Att-BERT.}
    \label{Fig:Att-BERT}
    \includegraphics[width=0.8\linewidth]{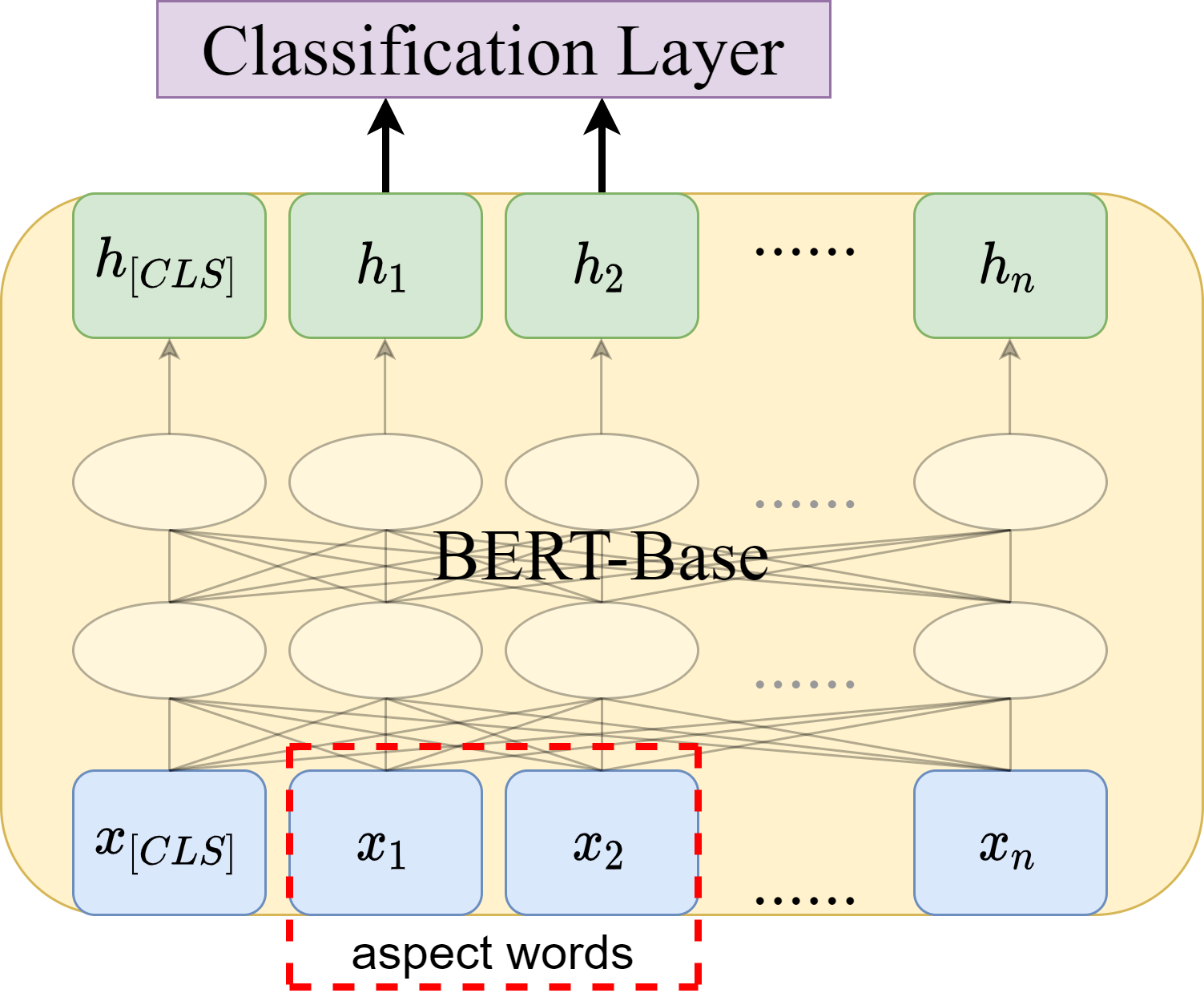}
	\caption{The illustration of BERTABSA.}
	\label{Fig:BERTABSA}
	\includegraphics[width=0.8\linewidth]{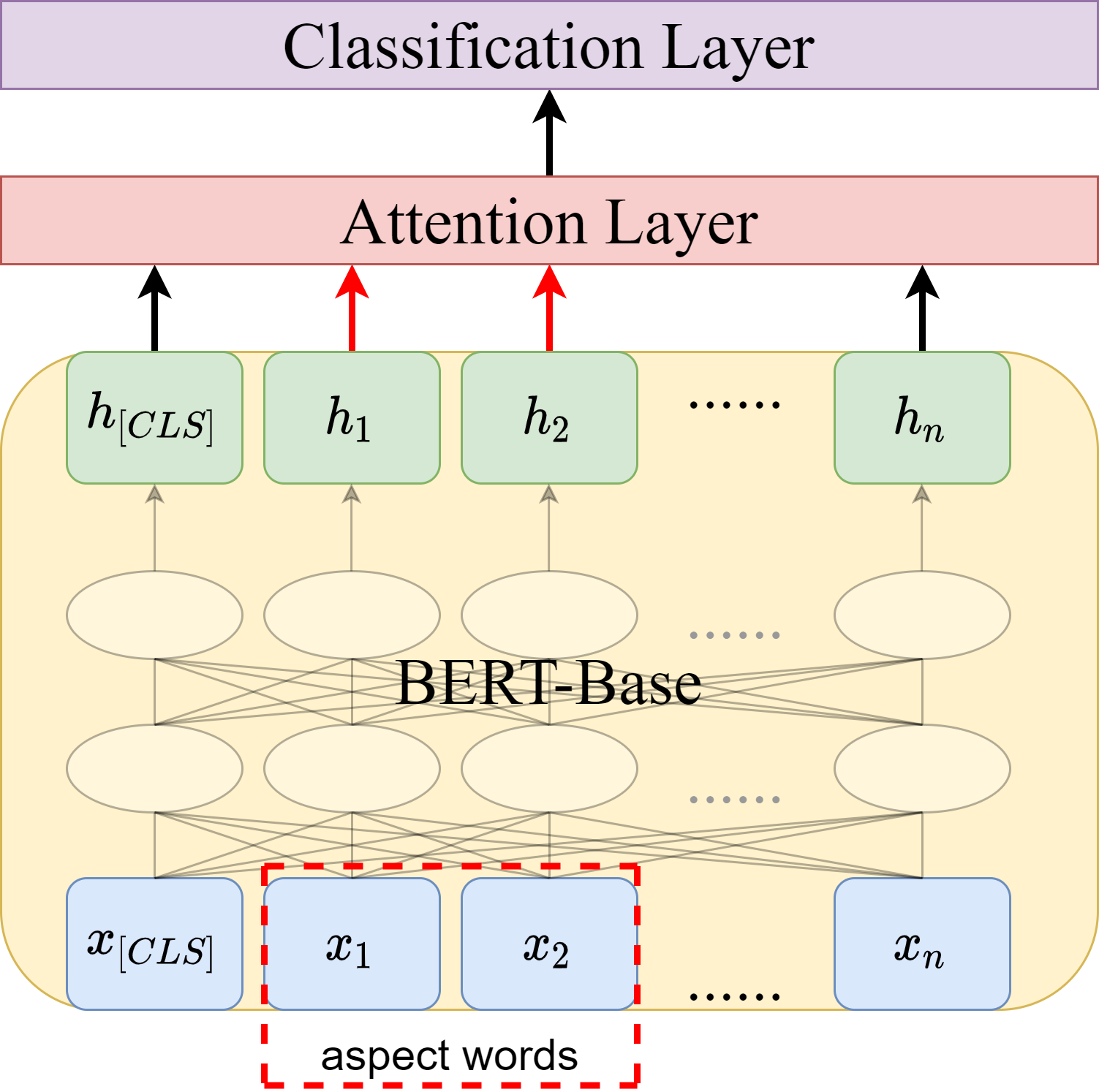}
	\caption{The illustration of Att-BERTABSA.}
	\label{Fig:Att-BERTABSA}
\end{figure}

Figure~\ref{Fig:Att-BERTABSA} shows the structure of Att-BERTABSA. Its structure is similar to Att-BERT, for adding a scale-dot attention layer after the output of BERT. However, different from Att-BERT, the hidden states of context words and aspect words are regarded as $Q$ and $K$ respectively and fed into the attention layer separately. To verify the effectiveness of our method, we make the same modifications on the Att-BERTABSA.

\end{appendices}

\end{document}